\documentclass[12pt,journal,compsoc]{IEEEtran}
\usepackage{cite}
\ifCLASSINFOpdf
  \usepackage[pdftex]{graphicx}
\else
  \usepackage[dvips]{graphicx}
\fi
\usepackage{amsfonts}
\usepackage{caption}
\usepackage{subcaption}
\usepackage{multirow}
 \usepackage[table,xcdraw]{xcolor}
\usepackage{booktabs}
\usepackage{tabularx}
\usepackage{listings}
\usepackage{lipsum}
\usepackage{xcolor}
\usepackage{stfloats}
\usepackage{ulem}
\usepackage{graphicx,wrapfig}
\usepackage[export]{adjustbox}

\usepackage[hyphens]{url}

\usepackage[T1]{fontenc}
\usepackage[utf8]{inputenc}
\usepackage{authblk}

\usepackage{soul}

\usepackage{array}
\makeatletter
\newcommand{\thickhline}{%
    \noalign {\ifnum 0=`}\fi \hrule height 1pt
    \futurelet \reserved@a \@xhline
}
\newcolumntype{"}{@{\hskip\tabcolsep\vrule width 1pt\hskip\tabcolsep}}
\makeatother

\usepackage{hyperref}
\hypersetup{
    colorlinks=true,
    citecolor=blue,
    urlcolor=cyan,
    linkcolor=red
}

\begin{document}

\title{\Large{Elementary Effects Analysis of factors controlling COVID-19 infections in computational simulation reveals the importance of Social Distancing and Mask Usage}}

\author[1,2]{Kelvin K. F. Li\thanks{kelvin.li@cuhk.edu.hk}}
\author[3]{Stephen A. Jarvis\thanks{s.a.jarvis@bham.ac.uk}}
\author[1]{Fayyaz Minhas\thanks{fayyaz.minhas@warwick.ac.uk}}
\affil[1]{Department of Computer Science, University of Warwick}
\affil[2]{Centre of Cyber Logistics, The Chinese University of Hong Kong}
\affil[3]{College of Engineering and Physical Sciences, University of Birmingham}

\renewcommand\Authands{ and }

\markboth{L\MakeLowercase{i et al.}, 2020}
{Shell \MakeLowercase{\textit{et al.}}: Bare Demo of IEEEtran.cls for Computer Society Journals}
\date{\normalsize\today}

\IEEEtitleabstractindextext{%
\begin{abstract}
COVID-19 was declared a pandemic by the World Health Organization (WHO) on March 11\textsuperscript{th}, 2020. With half of the world's countries in lockdown as of April due to this pandemic, monitoring and understanding the spread of the virus and infection rates and how these factors relate to behavioural and societal parameters is crucial for effective policy making. This paper aims to investigate the effectiveness of masks, social distancing, lockdown and self-isolation for reducing the spread of SARS-CoV-2 infections. Our findings based on agent-based simulation modelling show that whilst requiring a lockdown is widely believed to be the most efficient method to quickly reduce infection numbers, the practice of social distancing and the usage of surgical masks can potentially be more effective than requiring a lockdown. Our multivariate analysis of simulation results using the Morris Elementary Effects Method suggests that if a sufficient proportion of the population wore surgical masks and followed social distancing regulations, then SARS-CoV-2 infections can be controlled without requiring a lockdown.
\end{abstract}}

\maketitle
\IEEEpeerreviewmaketitle
\IEEEdisplaynontitleabstractindextext

\begin{figure*}[!t]
    \includegraphics[scale=1,center]{{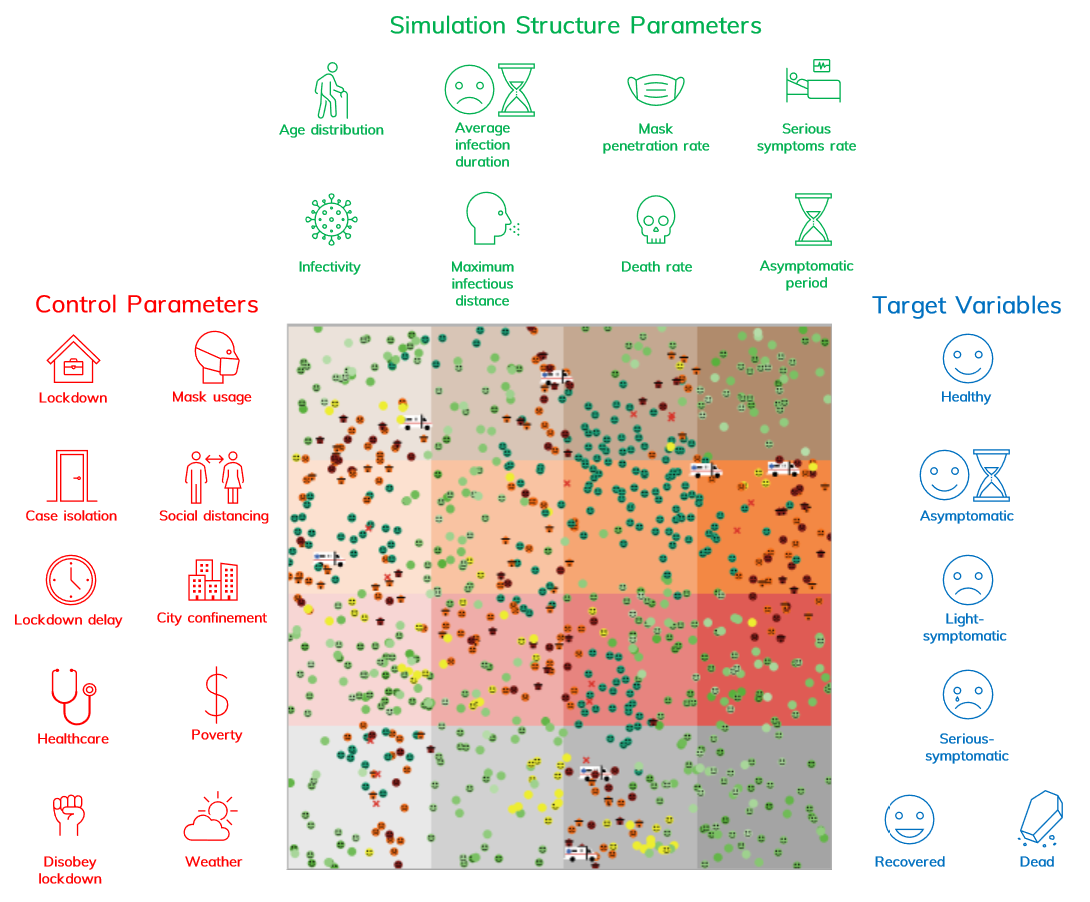}}
    \caption{Concept diagram of the Agent-based modelling framework, based on 3 types of parameters - Control Parameters, Simulation Structure Parameters and Target Variables}
    \label{bigfigure}
\end{figure*}

\section{Introduction}
\IEEEPARstart{C}{ovid-19}, informally known as the coronavirus, is a respiratory illness that is caused by the Severe Acute Respiratory Syndrome (SARS) Coronavirus 2 (SARS-CoV-2) \cite{Gorbalenya}. It was declared a pandemic by the World Health Organisation (WHO) on 11\textsuperscript{th} March 2020. As a vaccine for this virus can only be expected to be developed and distributed in mid-2021 \cite{ozkan}, controlling COVID-19's infection and fatality rates over the next year will be critically dependent on the successful implementation of public health measures, such as social distancing, isolation, quarantine and usage of masks \cite{anpan}.

It has almost become common sense that isolating symptomatic cases, enforcing a lockdown, social distancing and the usage of masks are amongst the most effective methods of reducing the spread of viral infections. Whilst there are many pre-existing univariate analyses evaluating the effectiveness of a particular preventative measure, such as the effectiveness of cloth masks against surgical masks in reducing COVID-19 infections \cite{maskresearch1} \cite{maskresearch2} and the effectiveness of social distancing \cite{socialdistancingresearch1} \cite{socialdistancingresearch2}, there is a lack of literature that reviews the effectiveness of these preventative measures relative to each other. For example, is social distancing more effective than the isolation of symptomatic cases? Is a lockdown the most effective option to reduce infection numbers? How useful are masks relative to the aforementioned strategies?

Traditionally, statistical and mathematical models, such as the Susceptible-Infected-Recovered (SIR) model, have been used to simulate and predict the trends of pandemics and answer some to the uncertainties. Unfortunately, they are mostly only useful at establishing the basic principles because of challenges such as numerical tractability and different research agendas. These models are usually oversimplified and fail to take into account human behaviour, human contact patterns, population diversity and variations of human hosts. This paper aims to answer the above questions via an agent-based modelling approach, which can efficiently capture emergent phenomena and inter-agent interactions, potentially providing a more accurate answer of the uncertainties that cannot be answered by traditional mathematical models.

\section{Methods}\label{materialmethod}
The following section describes the methods used to investigate the effectiveness of masks, social distancing, lockdown and self-isolation for reducing the spread of COVID-19 through the agent-based simulation framework developed in this work.

\subsection{Agent-Based Simulation}

In this work, we have developed an agent-based simulation based on the Susceptible-Infected-Recovered-Dead (SIRD) model by Li et al. \cite{LiSimulation}. 

Agent-based modelling (ABM) is a method of simulating the behaviour and interactions of autonomous agents in a particular environment across time steps. Autonomous agents can be thought of as individual entities that carry out some set of operations on behalf of a user but without the interference of that ownership entity \cite{agentbaseddef}. In this simulation, agents within the ABM are heterogeneous, meaning that each agent possesses a unique set of characteristics, including age group, gender, infection status, etc., that determine the agent's behaviour and the way it interacts with the other agents in the environment \cite{mongoose}. 

Agent-based models serve as an effective main decision support tool to impact the local and global impact of an upcoming pandemic and to test whether certain strategies would work against the pandemic \cite{agentpand}. They have been used to test combinations of individual-level (e.g., usage of masks) and mass-action strategies (e.g., lockdown) in different countries \cite{agentcountry}, as well as in smaller urban or rural areas \cite{agenturban}.

Since the 1990s, as more agent-based simulation packages and platforms such as Swarm, NetLogo, RePast and AnyLogic were released \cite{agentrelease}, agent-based modelling became increasingly used in Biology, to analyse the spread of epidemics \cite{agentbiology}.

In this section, we present a summary of the operational logic of the simulation. For a more detailed description, the interested reader is referred to our NetLogo codebase \cite{Netlogo}. This can be found in the following link:
\href{https://github.com/kelvin1338/NetLogo}{https://github.com/kelvin1338/NetLogo}\\

The model used in this paper consists of $N=10000$ individuals inside a $100 \times 100$ patch grid which models a set of $C=16$ cities as grid blocks. As increasing the population size will increase the model runtime, we chose a population size of 10000 because it is sufficiently large to provide good statistical robustness as the variances of the measured effects at the output of the model are fairly small across different simulation runs. As shown in Figure \ref{bigfigure}, at a parametric level, the simulation can be divided into three types of parameters: control variables, simulation structure parameters and target variables. The simulation control variables constitute the set of independent variables such as mask usage, social distancing, lockdown delay, case isolation/quarantine, etc., that can be changed to study their impact on target variables such as peak infection rate, number of hospitalizations and deaths, etc. Simulation structure parameters control how the simulation is run and includes factors such as the maximum number of time steps, mask effectiveness, symptomatic death rate, healthcare system capacity, etc. These variable settings determine the states of different agents in the simulation, thus controlling their behaviour. 

Initially, all individuals in the simulation are healthy except for the preset proportion of asymptomatic infected individuals. As these individuals move around, virus transmission to uninfected individuals occurs. Over time, the severity level of infected individuals can increase in a probabilistic manner which can lead to death or hospitalization. Individuals with serious symptoms can die based on a probability and the remaining symptomatic cases can recover. 

Below, we discuss different aspects of the agent-based simulation.

\subsubsection{States of Individuals}
Each individual in the simulation model has an associated set of state parameters: health (healthy, asymptomatic, light-symptomatic, serious-symptomatic, recovered and dead), mask usage (yes or no), quarantine status (upon developing symptoms), and hospitalization status (dependent on healthcare load and availability). The behaviour of the individual changes based on their state, which changes over time. Their state may also change at the next time step.

\subsubsection{Virus Transmission and Effect Modeling}
The simulation models the average infection duration, maximum infectious distance, age-wise disease severity and fatality rates, as well as the average number of days before asymptomatic individuals begin to be symptomatic. These factors affect how infections are transmitted in the population and can be changed by the user. However, they are determined based on published sources, as presented in table \ref{controlparameterstable}, and are kept fixed in this analysis. 

\subsubsection{Healthcare}

The simulation models a healthcare system where hospitalised individuals have reduced death rates and decreased recovery times. Individuals with severe symptoms are hospitalized based on a uniform probability distribution if the healthcare system has capacity.

\subsubsection{Modelling Movement and Lockdown}
In order to model the movement of individuals, all alive individuals who feel no symptoms can move randomly up to 0.2 units distance per tick. The user can initiate a lockdown at any time during the simulation, where everyone will stop moving except for a proportion of non-compliant individuals, which can also be specified by the user. A city-level lockdown can also be implemented which confines individuals to move only inside their own city.

\subsubsection{Modelling Mask Usage}
A clinical experiment conducted by MacIntyre et al. over 1607 people showed that particle penetration rate was 97\% through a cloth mask and 44\% through medical masks \cite{maskstat}. The experiments in this paper assume the use of surgical masks and hence the mask penetration rate used in the model is 44\%. However, this can be adjusted by the user, as well as the number of mask users. These mask users have their probability of getting infected and the probability of infecting others is reduced down to (mask penetration rate)$ \%$, as the remaining virus particles are 'filtered' by their mask.

If an infected mask user meets an uninfected mask user, then the virus particles are modelled as having to travel through two layers of masks; one from the infected person, and one from the uninfected person. Hence, in this particular situation, the probability of infecting others is reduced down to (mask penetration rate\%)$\textsuperscript{2}$.

\subsection{Simulation parameters}\label{settingssection}
A breakdown of different model parameters are provided in table \ref{controlparameterstable}. 

\begin{table*}[t]
\centering
\small
\begin{tabular}{|>{\arraybackslash}p{2.2cm}|>{\arraybackslash}p{2.6cm}|>{\arraybackslash}p{10.5cm}|}
\hline
\multicolumn{1}{|c|}{\textbf{Parameter}} & \multicolumn{1}{c|}{\textbf{Value}} & \multicolumn{1}{c|}{\textbf{Source / Justification}}                                       \\ \hline
Population size                          & 10000          & Relatively large sample to reduce anomalies                                                \\ \hline
\vtop{\hbox{\strut Population}\hbox{\strut Distribution}} &
  1000 people per age group &
  An additional 1000 people has been allocated to the middle 40-49 age group so that the total population is 10000 for the ease of comparison and interpretation. \\ \hline
\vtop{\hbox{\strut Size of}\hbox{\strut simulation} \hbox{\strut space}} &
  $100 \times 100$ patches &
  Not too spacious or densely populated and kept constant for the ease of comparison. Adjusted by setting max-pxcor and max-pycor to 100. \\ \hline
\vtop{\hbox{\strut Metres per}\hbox{\strut patch}}                         & 40 metres      & Arbitrary value for the ease of comparison                                                     \\ \hline
Total infection duration                 & 21 days        & Median infection duration for COVID-19 is 20.8 days (Bi et al., \cite{qifangbi})            \\ \hline
Asymptomatic period                      & 6 days         & Average asymptomatic period is 6 days (World Health Organisation., \cite{WHO73})            \\ \hline
Maximum infectious distance              & 2 metres       & Rough estimation based on advice from WHO and a study by Setti et al. \cite{2metresenough} \\ \hline
Infectivity                              & 100            & Infectivity should always be arbitrarily set to 100 for the ease of comparison                 \\ \hline
\vtop{\hbox{\strut Weather}\hbox{\strut conditions}}                       & Cold + Dry     & Arbitrary value for ease of comparison                                                     \\ \hline
\vtop{\hbox{\strut Mask usage*}\hbox{\strut Lockdown delay*} \hbox{\strut \% Ignore lockdown} \hbox{\strut Social distancing*} \hbox{\strut Healthcare}} &
  0 &
  All safety measures are disabled. \newline *Only enabled if the parameter of concern is being investigated by the sensitivity analysis experiment \\ \hline
\vtop{\hbox{\strut Enable lockdown*}\hbox{\strut City confinement}} &
  False &
  All safety measures are disabled. \newline *Only enabled if the parameter of concern is being investigated by the sensitivity analysis experiment \\ \hline
\end{tabular}
\caption{Control Variables}
\label{controlparameterstable}
\end{table*}

\subsubsection{Serious Symptomatic and Death Rates}
The rate of an infected individual developing serious symptoms was obtained by adjusting the hospitalisation rates from a study by Verity et al. \cite{verity} to match expected rates in the highest age group. This is shown in the second column of table \ref{deathstats}.

As the simulation model assumes that only individuals with serious symptoms are subject to death, we calculated the adjusted fatality rate for each age group, used for our simulation model, based on the real-life infection fatality ratio (IFR) in the 3\textsuperscript{rd} column of table \ref{deathstats}, which was also obtained from a study by Verity et al. \cite{verity}. The aforementioned adjusted fatality rate is presented in the 4\textsuperscript{th} column of table \ref{deathstats}, and is simply the third column divided by the second column, multiplied by 100.

\begin{table}[!t]
\footnotesize
\begin{center}
\begin{tabular}{|>{\centering}p{0.9cm}|>{\centering\arraybackslash}p{1.85cm}|>{\centering\arraybackslash}p{2cm}|>{\centering\arraybackslash}p{2cm}|}
\hline
Age group & Serious symptoms rate (\%) [SR] \cite{neil} \cite{verity} & Real infection fatality ratio (\%) [IFR] \cite{neil} \cite{verity} & Model Fatality rate (\%) [100$\times$IFR/SR] \\ \hline
0-9   & 0.1  & 0.002 & 2.00 \\ \hline
10-19 & 0.3  & 0.006    & 2.00 \\ \hline
20-29 & 1.2  & 0.03    & 2.50  \\ \hline
30-39 & 3.2  & 0.08    & 2.50  \\ \hline
40-49 & 4.9  & 0.15  & 3.06  \\ \hline
50-59 & 10.2 & 0.60 & 5.88  \\ \hline
60-69 & 16.6 & 2.2 & 13.25  \\ \hline
70-79 & 24.3 & 5.1 & 20.99   \\ \hline
80+   & 27.3 & 9.3 & 34.07   \\ \hline
\end{tabular}
\caption{Symptomatic and death rates}
\label{deathstats}
\end{center}
\end{table}

\subsubsection{Ranges of Analysis Variables}
Recall that in this research paper, the four main independent variables of interest are mask usage, social distancing, lockdown and isolation of symptomatic cases.

For mask usage and isolation rate, the minimum and maximum values are the cases where nobody follows the rule, and everybody follows the rule, respectively. Hence, 0 is the minimum, and 100 is the maximum.

For social distancing, the minimum value is the case where nobody follows social distancing rules, which is 0. The maximum value was chosen to be 2.5 because realistically most countries enforce social distancing up to 2 metres. Also, the maximum radius of infection is estimated to be approximately 2 metres.

For lockdown delay, the minimum bound was chosen to be 7 days because it would be unrealistic for a government to enforce a lockdown immediately from the moment the first person in the country catches the infection. The maximum bound was chosen to be 32 days because from the visual analysis and univariate analysis, delaying the lockdown beyond 20-24 days has a negligible effect on the peak of the infection curve.

\begin{table}[!tb]
\scriptsize
\begin{tabular}{l|>{\centering\arraybackslash}p{1.35cm}|>{\centering\arraybackslash}p{1.1cm}|>{\centering\arraybackslash}p{1.2cm}|>{\centering\arraybackslash}p{1.7cm}|}
\cline{2-5}
                                             & \multicolumn{4}{c|}{\textbf{Parameter}} \\ \cline{2-5} 
 & \textbf{Social distancing metres } & \textbf{Mask usage rate} & \textbf{Lockdown delay} & \textbf{Symptomatic Isolation rate} \\ \hline
\multicolumn{1}{|>{\centering\arraybackslash}p{1.3cm}|}{\textbf{Maximum}} & 2.5      & 100      & 32      & 100     \\ \hline
\multicolumn{1}{|>{\centering\arraybackslash}p{1.3cm}|}{\textbf{Minimum}} & 0        & 0        & 7        & 0      \\ \hline
\end{tabular}
\caption{Boundaries of independent variables}
\label{EEMrange}
\end{table}

\subsubsection{Target Variables}
From each simulation run, the number of infections is monitored. As the simulation involves stochasticity, the peak percentage of infections across multiple simulation runs for a given parameter setting is used as a simple and interpretable variable of interest. 

For each configuration of parameters, a more in-depth analysis can be provided, allowing the \% of the population infected at any given time to be recorded, as well as other more specific target variables such as the asymptomatic, light-symptomatic, serious-symptomatic, recovered and dead populations at any point in time.

\subsection{Sensitivity Analysis}
We are interested in studying the impact of each of the independent variables on the dependent variables, as long as the virus-related simulation parameters are kept fixed. For this purpose, we use univariate analysis and the method of elementary effects as discussed below. 

\subsubsection{Univariate Analysis}\label{univariatedebrief}
In a univariate sensitivity analysis, a particular parameter is adjusted, whilst the remaining parameters remain unadjusted and kept constant. The peak \% of daily infections are recorded for each run, i.e., the peak point of the infection curve. This was conducted for each of the four independent variables in increments of 20\% from the absolute minimum to the absolute maximum. To ensure statistical robustness in our results, the simulation is run 12 times for each increment of the independent variable. The results are then analysed and plotted on a scatter graph.

The sensitivity index of each parameter's median peak infections were calculated using the corresponding formula below and was compared to the other parameters. 
\small
\begin{equation}
sensitivity = \frac{Y_{max} - Y_{min}}{Y_{max}}
\end{equation}
\normalsize

Here, $Y_{max}$ and $Y_{min}$ correspond to the maximum and minimum total infections respectively, for each set of parameter settings.

Sensitivity index was chosen because it is an easily interpretable metric that summarises how much an NPI can affect infection numbers across all simulation runs. As each set of parameter configurations is run 12 times and 5 increments are analysed for each of the four NPIs, this meant that there are 60 simulation runs for each non-pharmaceutical intervention, and 240 simulation runs in total.

An experiment is run for each of the four independent parameters according to the range specified in table \ref{EEMrange}.

\subsubsection{Elementary Effects Method}
Whilst the previous univariate sensitivity analysis provided insight regarding the strength and effect of each individual variable, it fails to capture multivariate correlation. The following section provides a multivariate analysis of the four main preventative measures variables - mask usage, social distancing, symptomatic isolation and lockdown delay, and how they interact with each other via Morris's Elementary Effects Method \cite{morris}, which is an adaptation of the One-At-a-Time (OAT) design approach that identifies the most important variables in determining the output using a small number of simulations. 

Given a process $Y = g(x_1, ..., x_k)$ dependent upon $k$ factors with $x \in [0,1]$, $r$ trajectories are generated by EEM, which is divided into a $k$-dimensional grid consisting of $\rho$ levels of equal size. A trajectory begins with randomly selected base values $x\mbox{*} = [x_1\mbox{*}, ... , x_k\mbox{*}]$ in the $\rho$-level grid. This base vector is used to generate $k$ different parameter vectors for a particular trajectory. For this study the EEM trajectories were generated by the SALib library \cite{salib} using a random seed number of 1.

The first parameter vector $x\textsuperscript{(1)}$ is generated by adding or subtracting $\Delta$ to one of the parameters. Similarly, $x\textsuperscript{(2)}$
is obtained by adding or subtracting $\Delta$ to another parameter. The parameters defined here are used in equations 2-6 below.

After obtaining the values for each parameter vector in the trajectory, the elementary effect of a parameter can then be obtained using the formula below:

\small
\begin{equation}
EE_i = \frac{Y(x_1, ..., x_i + \Delta, ..., x_k) - Y(x_1, ..., x_i, ... , x_k)}{\Delta}
\end{equation}
\normalsize

After the elementary effects were obtained, the corresponding mean, absolute-mean and variance of the elementary effect of a variable was calculated using the three formulas below respectively:

\begin{equation}
\mu_i = \frac{1}{r}\sum\limits_{j=1}^{r} EE_i^j 
\end{equation}

\begin{equation}
\mu_i\mbox{*} = \frac{1}{r}\sum\limits_{j=1}^{r} |EE_i^j|
\end{equation}

\begin{equation}
\sigma_i = \frac{1}{r-1}\sum\limits_{j=1}^{r} (EE_i^j - \mu_i)^2 
\end{equation}

The reason $\mu_i\mbox{*}$ is introduced and uses absolute values is to prevent certain elementary effects from cancelling out due to opposite signs.
Note that a high value of $\mu_i\mbox{*}$ implies that the output is very sensitive to the value of parameter $i$. A large value of $\sigma$ indicates that there is an interaction between this parameter and other parameters and that the parameter is interconnected to the values of other parameters.

After calculating the corresponding values for a particular variable, its 'rank' can be calculated using the following formula, which determines the strength of the parameter's effect relative to other parameters:

\begin{equation}
Rank_i = \sqrt{\mu_i\mbox{*}^2 + \sigma_i}
\end{equation}

Finally, the results of the elementary effects analysis are compared to the real-life trends of COVID-19 in the United Kingdom, Hong Kong and Italy, which was chosen because of their contrasting ways of dealing with COVID-19. The information is obtained from various medical and news sources, as well as interviews and questionnaires. The information gathered from the three countries is summarised and placed in table \ref{informationtable}.

\begin{table*}[!t]
\centering
\footnotesize
\setlength\tabcolsep{3pt}
\setlength\extrarowheight{2pt}
\begin{tabular}{|l|l|l|l|}
\hline
                                                  & \multicolumn{1}{c|}{United Kingdom}                       & \multicolumn{1}{c|}{Hong Kong}                           & \multicolumn{1}{c|}{Italy}                    \\ \hline
Date of first case                                & 23\textsuperscript{rd} January \cite{novelUK} & 22\textsuperscript{nd} January \cite{HKindexcase} & 31\textsuperscript{st} January \cite{ITindex} \\ \hline
Median age                                        & 40.5 \cite{UKdistribution}                    & 44.8 \cite{HKdistribution}                        & 47.3 \cite{ITdistribution}                    \\ \hline
Proportion of people aged 65+                     & 18.48\% \cite{UKdistribution}                 & 18.48\% \cite{HKdistribution}                     & 22.08\% \cite{ITdistribution}                 \\ \hline
Population                                        & 67.78 million \cite{popdensity}               & 7.48 million \cite{popdensity}                    & 60.48 million \cite{popdensity}               \\ \hline
Population Density (People/km\textsuperscript{2}) & 281 \cite{popdensity}                         & 7140 \cite{popdensity}                            & 206 \cite{popdensity}                         \\ \hline
Urban population                                  & 83\% \cite{popdensity}                        & 100\% \cite{popdensity}                           & 69.5\% \cite{popdensity}                      \\ \hline
Rural population                                  & 17\% \cite{popdensity}                        & 0\% \cite{popdensity}                             & 30.5\% \cite{popdensity}                      \\ \hline
Previous experience with similar outbreaks?       & No                                            & Yes (SARS 2003) \cite{HKsars}                     & No                                            \\ \hline
\% reported to follow lockdown rules              & 89\% \cite{UKsurvey1}                         & N/A                                               & N/A                                           \\ \hline
Lockdown implemented?                             & Yes \cite{UKlaw2}                             & No                                                & Yes \cite{ITlockdown1}                        \\ \hline
14-day quarantine implemented?                    & Yes (Not properly enforced)                   & Yes (Strict) \cite{HKborderclose}                 & No                                            \\ \hline
Usage of Masks                                    & Rare                                          & Strictly followed from the start                  & Mandatory after some time                     \\ \hline
Main source of healthcare                         & NHS \cite{nhsuk}                                          & Public + Private healthcare                       & SSN \cite{IThealthcare}                                           \\ \hline
Healthcare free for citizens?                     & Yes \cite{nhsuk}                                           & No \cite{HKprivate}                                                & Yes   \cite{IThealthcare}                                     \\ \hline
\% of citizens who can afford healthcare          & 100\% \cite{nhsuk}                                         & 92\% \cite{HKprivate}                             & 100\% \cite{IThealthcare}                                        \\ \hline
Total doctors                                     & 280000 \cite{guardiandoctor}                  & 14290 \cite{HKdoctornumber}                       & 427213 \cite{ITdoctors}                       \\ \hline
Doctors per 1000 people                           & 2.8 \cite{guardiandoctor}                     & 1.9 \cite{HKdoctornumber}                         & 4 \cite{ITdoctors2}                           \\ \hline
Total nurses                                      & 661000 \cite{UKnurse}                         & 56723 \cite{HKnursenumber}                        & 418461 \cite{ITdoctors3}                      \\ \hline
Total nurses per 1000 people                      & 8.17 \cite{UKnurse}                           & 7.3 \cite{HKnursenumber}                          & 5.74 \cite{ITdoctors3}                        \\ \hline
Hospital beds per 1000 people                     & 6.6 \cite{UKbedICU}                           & 7.1 \cite{HKbeds}                                 & 3.4 \cite{ITdoctors3}                         \\ \hline
\end{tabular}
\captionof{table}{Summarised Background Research of UK, Hong Kong and Italy}
\label{informationtable}
\end{table*}

\subsubsection*{Implementation of EEM}
The values for each of the independent variables were standardised to a range between 0 and 1. The value of $\Delta$ was chosen to be 0.2, which meant that each parameter can take 6 unique standardised values - 0, 0.2, 0.4, 0.6, 0.8 or 1.0. As there are four parameters, it follows that there are $6^4 = 1296$ possible permutations. Using BehaviorSpace, 1296 simulations were run, once for each permutation. The peak infection \% was recorded for each run.

30 arbitrary trajectories were generated with SALib, a Python library that contains a Morris-Elementary-Effect toolkit. From the definition of EEM, this meant that there are 5 simulations associated with each trajectory, meaning $5 \times 30 = 150$ different simulations were required. Recall earlier that the peak infection \% for all 1296 permutations are available. Hence, for each permutation, the peak \% of infections was extracted from the list of 1296 runs and mapped accordingly.

Now that there is a list of trajectories with the corresponding outcome, the final step was to calculate $\mu, \mu\mbox{*}, \sigma$ and the rank for each parameter. Using the 'analyze' function in SALib, the $\mu, \mu\mbox{*}$ and $\sigma$ was computed, and the rank was manually calculated afterwards. The results are presented later in section \ref{EEMresultssection}.

\subsection{Implementation}
The interactive agent-based simulation model has been uploaded to GitHub along with detailed usage instructions, which can be found on the following link:
\noindent\href{https://github.com/kelvin1338/NetLogo}{https://github.com/kelvin1338/NetLogo}

\begin{table}[b]
\centering
\scriptsize
\begin{tabular}{c|>{\centering\arraybackslash}p{1.4cm}|>{\centering\arraybackslash}p{0.8cm}|>{\centering\arraybackslash}p{1.2cm}|>{\centering\arraybackslash}p{1.7cm}|}
\cline{2-5}
\multicolumn{1}{l|}{}                          & \multicolumn{4}{c|}{\textbf{Parameter (i)}} \\ \cline{2-5} 
\multicolumn{1}{l|}{} & \textbf{Social distancing metres} & \textbf{Mask usage rate} & \textbf{Lockdown delay} & \textbf{Symptomatic Isolation rate} \\ \hline
\multicolumn{1}{|c|}{\textbf{\vtop{\hbox{\strut Sensitivity}\hbox{\strut index}}}} & 0.784 & 0.692 & 0.683 & 0.238    \\ \hline
\end{tabular}
\caption{Sensitivity index of the four dependent variables}
\label{UNIresults}
\end{table}

\section{Results}
The following section provides a breakdown of the results of all experiments proposed in chapter \ref{materialmethod}.

\subsection{Univariate Analysis Results}\label{sens}
The following subsection provides an analysis for each of the four independent variables. From the results presented in table \ref{UNIresults}, it was observed that social distancing has a very high sensitivity index, whilst the mask usage rate and lockdown delay both have relatively high and similar sensitivity index. Isolating symptomatic cases results in the lowest sensitivity index. It was also observed that a lockdown reduces the infection numbers to 0 in the shortest amount of time, whilst social distancing and mask usage are relatively slower. Both of these measures flatten the curve as the respective measure is followed by more people, although it takes longer for the cases to reach 0 again. Purely isolating symptomatic cases reduces the peak, although there is little impact on the distribution of the infection curve. Hence, this has the last significant effect out of the four safety measures.

\begin{table*}[t]
\setlength{\tabcolsep}{1.8pt}
\centering
\scriptsize
\hspace*{-1cm}
\begin{tabular}{|>{\centering}p{1.35cm}"c|c|c|c|c|c"c|c|c|c|c|c"c|c|c|c|c|c"c|c|c|c|c|c|} 
\hline
\multicolumn{1}{|l"}{} & \multicolumn{6}{c"}{\textbf{Social Distancing (metres) }} & \multicolumn{6}{c"}{\textbf{Mask Usage rate (\%)}} & \multicolumn{6}{c"}{\textbf{Lockdown delay (days) }} & \multicolumn{6}{c|}{\textbf{Symptomatic Case Isolation (\%) }}  \\ 
\hline
                       & 0     & 0.5   & 1     & 1.5   & 2     & 2.5               & 0     & 20    & 40    & 60    & 80    & 100        & 7     & 12    & 17    & 22    & 27    & 32           & 0     & 20    & 40    & 60    & 80    & 100                     \\ 
\hline
Median                 & 45.38 & 26.94 & 26.09 & 18.38 & 12.75 & 9.79              & 45.59 & 33.10 & 25.87 & 19.20 & 14.06 & 9.68       & 14.45 & 25.09 & 32.54 & 40.53 & 42.12 & 45.52        & 44.13 & 40.53 & 41.34 & 39.29 & 36.76 & 33.64                   \\ 
\hline
Mean                   & 44.27 & 26.91 & 25.69 & 18.53 & 13.08 & 10.06             & 44.63 & 33.87 & 25.61 & 18.95 & 14.09 & 9.85       & 14.29 & 25.03 & 32.33 & 40.90 & 41.97 & 44.63        & 43.19 & 41.39 & 41.76 & 38.96 & 37.48 & 33.77                   \\ 
\hline
Range                  & 6.69  & 8.03  & 5.30  & 3.54  & 4.45  & 3.22              & 9.38  & 6.72  & 7.18  & 3.77  & 3.79  & 3.35       & 4.51  & 4.94  & 8.29  & 8.76  & 8.62  & 11.30        & 8.63  & 12.40 & 13.18 & 11.19 & 14.12 & 9.13                    \\ 
\hline
Variance               & 4.98  & 4.49  & 3.10  & 2.32  & 1.78  & 1.12              & 8.16  & 5.52  & 3.36  & 1.69  & 1.38  & 0.97       & 1.52  & 2.30  & 5.85  & 6.65  & 4.43  & 11.62        & 8.16  & 11.75 & 12.34 & 11.91 & 16.16 & 8.44                    \\ 
\hline
Standard Deviation     & 2.23  & 2.12  & 1.76  & 1.52  & 1.33  & 1.06              & 2.86  & 2.35  & 1.83  & 1.30  & 1.18  & 0.98       & 1.23  & 1.52  & 2.42  & 2.58  & 2.10  & 3.41         & 2.86  & 3.43  & 3.51  & 3.45  & 4.02  & 2.91                    \\ 
\hline
Standard Error         & 0.19  & 0.18  & 0.15  & 0.13  & 0.11  & 0.09              & 0.24  & 0.20  & 0.15  & 0.11  & 0.10  & 0.08       & 0.10  & 0.13  & 0.20  & 0.21  & 0.18  & 0.28         & 0.24  & 0.29  & 0.29  & 0.29  & 0.33  & 0.24                    \\
\hline
\end{tabular}
\caption{Summary statistics for social distancing, mask usage, lockdown delay and symptomatic case isolation, measuring \% peak of active cases}
\label{resultstable}
\end{table*}

\subsubsection{Social Distancing}
Social distancing has a sensitivity index of 0.784, based on the results presented in table \ref{resultstable}, which is the highest out of the four parameters. From Figure \ref{socialdistancingfigure12}, it can be seen that there is a significant decrease in the severity of infections as soon as social distancing was enforced. As the distance of social distancing further increased, the severity of infections decreased, following a negative logarithmic trend. Figure \ref{socialdistancingfigure1} shows the infection curve corresponding to the number of active cases at a given time. As the social distancing metres increased, the peak number of active cases decreased, but so did the rate of decrease in cases. This is known as flattening the curve \cite{flatcurve}.

\subsubsection{Mask Usage Rate}
\begin{figure*}[t]
\begin{subfigure}{.5\textwidth}
  \centering
  \includegraphics[width=.8\linewidth]{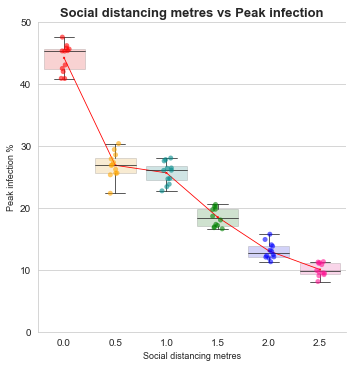}  
  \caption{Social distancing results (12 runs per setting)}
  \label{socialdistancingfigure12}
\end{subfigure}
\begin{subfigure}{.5\textwidth}
  \centering
  \includegraphics[width=.8\linewidth]{{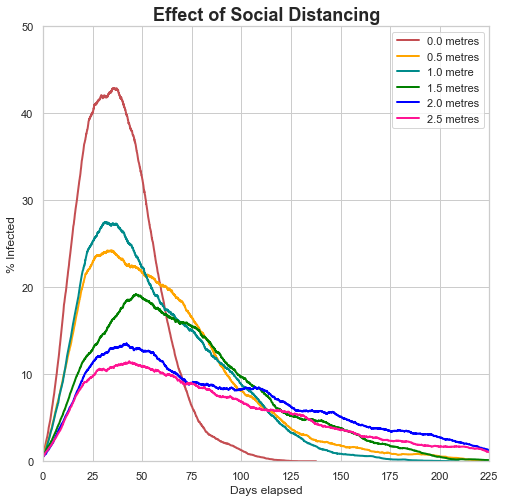}}  
  \caption{Social distancing results (1 run per setting)}
  \label{socialdistancingfigure1}
\end{subfigure}

~\\
~\\

\begin{subfigure}{.5\textwidth}
  \centering
  \includegraphics[width=.8\linewidth]{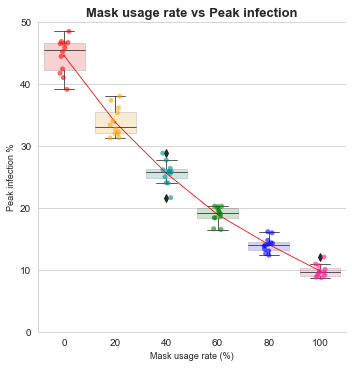}  
  \caption{Mask usage results (12 runs per setting)}
  \label{maskfigure12}
\end{subfigure}
\begin{subfigure}{.5\textwidth}
  \centering
  \includegraphics[width=.8\linewidth]{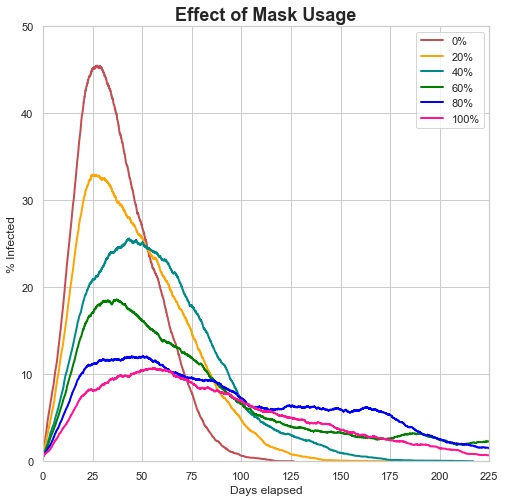} 
  \caption{Mask usage results (1 run per setting)}
  \label{maskfigure1}
\end{subfigure}
\caption{Univariate analysis results of social distancing and mask usage}
\label{sdmask}
\end{figure*}

Mask usage has a sensitivity index of 0.692, based on the results presented in table \ref{resultstable}, which is almost identical to lockdown delay but less than social distancing. From Figure \ref{maskfigure12}, it was noticed that as the mask usage rate increased from 0 to 100, the severity of the virus decreases with a negative-logarithmic trend, although it is more linear compared to social distancing. Figure \ref{maskfigure1} shows that as mask usage increases, the peak of the curve flattens whilst taking longer to reach 0 cases again for all configurations.

\subsubsection{Lockdown Delay}
Lockdown delay has a sensitivity index of 0.683, based on the results from table \ref{resultstable}, which indicates that it appears to have almost as much effect as mask usage but less than social distancing. From Figure \ref{lockdowndelayfigure12}, the severity of the virus increased almost linearly and very steeply as the lockdown delay was increased until a certain point, which highlights the importance of enforcing a lockdown early before the cases become out of control. In addition, it can be seen from Figure \ref{lockdowndelayfigure1} that enforcing a lockdown after the peak infection has been reached has very little effect on the severity of infections.

\subsubsection{Symptomatic Case Isolation Rate}
\begin{figure*}[t]
\begin{subfigure}{.5\textwidth}
  \centering
  \includegraphics[width=.8\linewidth]{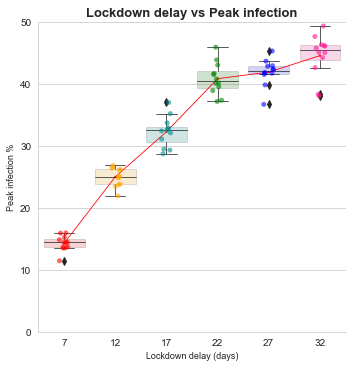}  
  \caption{Lockdown delay results (12 runs per setting)}
  \label{lockdowndelayfigure12}
\end{subfigure}
\begin{subfigure}{.5\textwidth}
  \centering
  \includegraphics[width=.8\linewidth]{{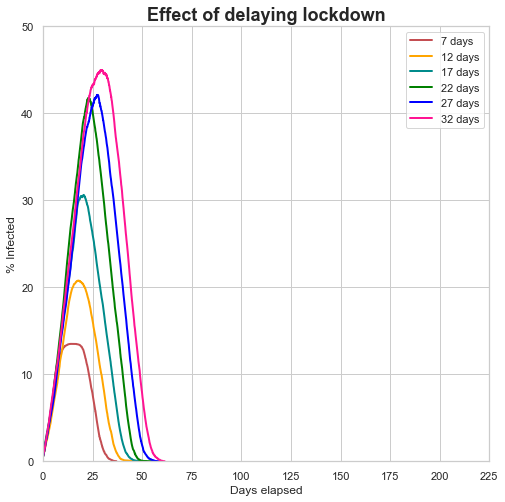}}  
  \caption{Lockdown delay results (1 run per setting)}
  \label{lockdowndelayfigure1}
\end{subfigure}

~\\
~\\

\begin{subfigure}{.5\textwidth}
  \centering
  \includegraphics[width=.8\linewidth]{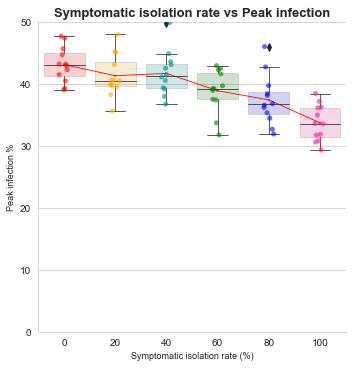}  
  \caption{Case isolation results (12 runs per setting)}
  \label{isolationfigure12}
\end{subfigure}
\begin{subfigure}{.5\textwidth}
  \centering
  \includegraphics[width=.8\linewidth]{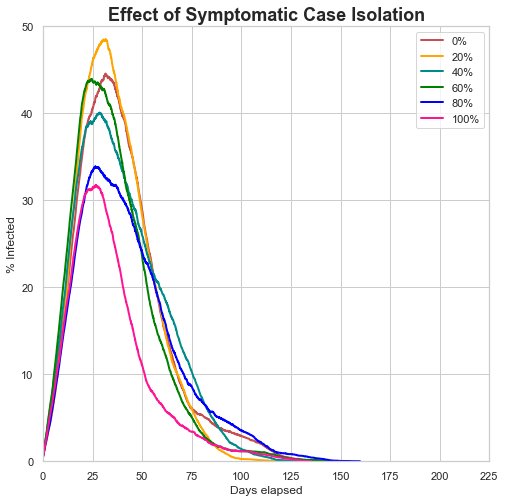} 
  \caption{Case isolation results (1 run per setting)}
  \label{isolationfigure1}
\end{subfigure}
\caption{Univariate analysis results of lockdown delay and symptomatic case isolation}
\label{ldiso}
\end{figure*}

Symptomatic case isolation has a sensitivity index of 0.238, based on the results from table \ref{resultstable}, which is the lowest of the four parameters, meaning it has the least effect out of the four safety measures. From Figure \ref{isolationfigure12}, it was observed that using this parameter alone will result in a high variance for all settings compared to the other three preventative measures, suggesting isolating symptomatic cases alone will not consistently contain the virus. Figure \ref{isolationfigure1} shows that isolating cases alone linearly reduces the peak of the infection curve, but it is relatively less effective in flattening the curve, as it takes a similar time to reach 0 cases again.

\subsection{Elementary Effects Analysis Results}\label{EEMresultssection}

The results of the elementary effects analysis are shown in table \ref{EEMresults}. From the results in this table, social distancing is the most effective measure in reducing the peak of the infection curve, followed by mask usage, lockdown delay and finally isolation of symptomatic cases. Social distancing has the highest value of $\mu_i\mbox{*}$, meaning the peak of the infection curve is very sensitive to the measure. It also has the highest value of $\sigma$, meaning that there is a large interaction between this parameter and the other three parameters.

As seen from the results table, enforcing a lockdown has a relatively lower  $\sigma$ and $\mu_i\mbox{*}$ compared to masks and social distancing, indicating that it has less effect on the peak of the infection curve than masks and social distancing regardless of parameter interactions. Whilst the univariate analysis earlier showed that a lockdown is the most effective measure at reducing the peak of the infection curve as quickly as possible, this experiment showed that a lockdown may not be necessary in the first place, as there would potentially be no significant infection peak initially if social distancing was followed and masks were used by everyone. 

\begin{table}[!t]
\begin{tabular}{c|>{\centering\arraybackslash}p{1.6cm}|>{\centering\arraybackslash}p{1cm}|>{\centering\arraybackslash}p{1.4cm}|>{\centering\arraybackslash}p{1.8cm}|}
\cline{2-5}
\multicolumn{1}{l|}{}                          & \multicolumn{4}{c|}{\textbf{Parameter (i)}} \\ \cline{2-5} 
\multicolumn{1}{l|}{} & \textbf{Social distancing metres} & \textbf{ Mask usage rate } & \textbf{Lockdown delay} & \textbf{Symptomatic Isolation rate} \\ \hline
\multicolumn{1}{|c|}{\textbf{$\mu_i\mbox{*}$}} & 4.275 & 4.039 & 2.014 & 1.377    \\ \hline
\multicolumn{1}{|c|}{\textbf{$\mu_i$}}         & -3.981 & -4.039 & 1.641 & -0.777    \\ \hline
\multicolumn{1}{|c|}{\textbf{$\sigma_i$}}      & 5.255  & 3.246  & 2.881 & 1.667    \\ \hline
\multicolumn{1}{|c|}{\textbf{$Rank_i$}}        & 4.850  & 4.422  & 2.634 & 1.888    \\ \hline
\end{tabular}
\caption{Results of elementary effects for 30 trajectories, $p = 6$ and $\Delta = 0.2$}
\label{EEMresults}
\end{table}

\subsection{Relationship of EEM Results with Trends in Different Countries}
In order to understand the correspondence of simulation results with real-world data, we have studied the trends of viral transmission across different countries. Specifically, we obtained the COVID-19 timelines, the main preventative measures and infection numbers over time for the United Kingdom, Hong Kong and Italy. We then compared these trends and numbers to the results in our Elementary Effects results in table \ref{EEMresults}.

The experiment results showed the importance of surgical mask usage; and strongly supports Hong Kong's strategy in successfully containing the virus, where they showed a strong emphasis on mask usage and social distancing from the very beginning \cite{HKfast}, whilst not enforcing any total lockdowns throughout the whole process. For example, some restaurants, entertainment venues and sports centres were closed and the majority of the population began wearing masks before the virus even reached Hong Kong.

\begin{figure}[!tb]
    \includegraphics[scale=0.30,center]{{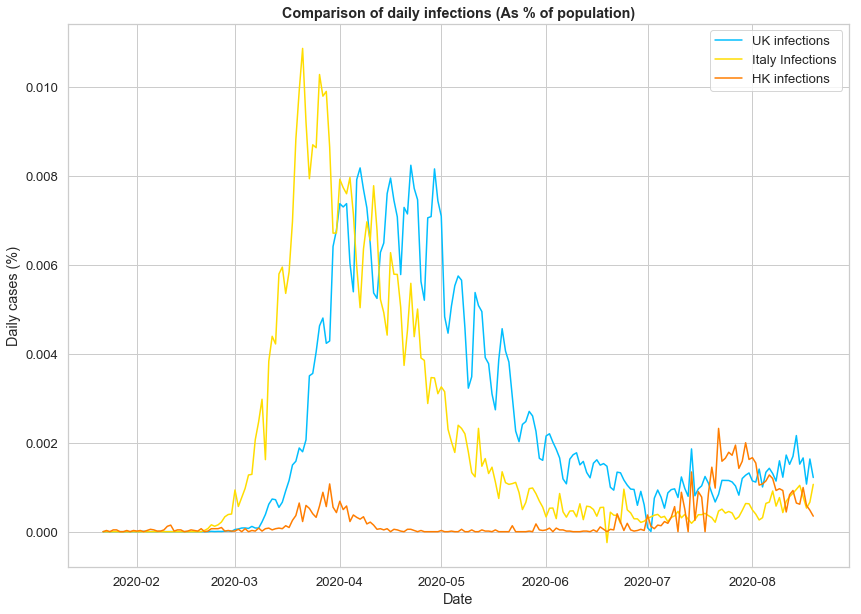}}
    \caption{Daily COVID-19 cases in the United Kingdom, Hong Kong and Italy \cite{johnshopkins}}
    \label{countriescomparison}
\end{figure}

The results from the EEM shows that the symptomatic isolation rate is the least effective safety measure out of the four variables, which can be explained by COVID-19's long asymptomatic period. This estimation is fairly relatable with the United Kingdom and Italy, who were heavily focused on the isolation of symptomatic cases and were much slower than Hong Kong to enforce any other safety measures \cite{UKtooslow}. For example, it took the United Kingdom longer than a month before restaurants, pubs, schools and entertainment venues were shut, and several more months before people started wearing masks. This lead to the infection curve rising to drastically higher values compared to Hong Kong as seen in Figure \ref{countriescomparison}, which compares the percentage of the overall population infected daily between the three countries. It can be seen that the cases in the UK and Italy were exponentially higher than in Hong Kong, and they were left with no choice but to implement a lockdown, which is forecasted by the EEM and univariate analysis results.

\subsection{Comparison of Model Infection Trends with Historical Trends}\label{comparisonsection}
We conducted a further investigation to test the validity of our model by comparing the trend of the infection numbers to real-world data. For this simulation, Hong Kong, Italy and the United Kingdom have been chosen due to their unique infection trends. The model simulation structure parameters were adjusted accordingly to resemble the population of the simulated country, whilst the control parameters are adjusted during the simulation, at different points, to loosely follow the historical timeline and actions taken by the countries. A simulation was run for each country up to the end of the first major wave of infections and the results are shown in the following subsections. 

Note that in real-life, many asymptomatic cases are not recorded, and hence the 'Symptomatic' blue curve in Figures \ref{hkfigure}, \ref{itfigure} and \ref{ukfigure} follow the reported historical figures most closely, as symptomatic cases are most likely detected and reported in real-life. The other curves provide a more detailed analysis and help us estimate the true number of cases.

In the following sections, we discuss the results for Hong Kong, Italy and the United Kingdom. For each country, the correlation between normalised time series of the predicted and actual number of cases are calculated, together with the corresponding p-values. Note that as the model results were measured hourly and historical results were measured daily, the model results were randomly sampled so that it is the same length as the historical data. This allows us to compute correlation coefficients.

\subsubsection{Hong Kong}
The age distribution used for the simulation of Hong Kong is provided in table \ref{HKpoptable}, calculated from data provided by the United Nations \cite{UNpopulation}.

As the first two waves of infections in Hong Kong contained very few infections, the simulation is run until the end of the first major spike of infections, known as the 'third wave', so that sufficient data can be compared.

\subsubsection*{Hong Kong Validation Setup}
Having experienced the SARS outbreak in 2003, a similar virus, it was established that Hong Kong took a more cautious approach compared to the UK and Italy in their COVID-19 strategy since the beginning \cite{bloomberghk}. Hence this model assumes 1.5 metres of social distancing and 99\% mask usage from day 0.

The model is set up to simulate Hong Kong from the middle of March, just as the first spike in cases was about to begin. Hong Kong severely tightened their borders very soon after the initial case \cite{HKborderclose}. Hence, this model assumed that it took just 7 days to tighten border controls. To model imported COVID-19 cases, 2 out of 10000 people are randomly infected every week from day 7.

Very soon after tightening borders, Hong Kong began enforcing a curfew, closing down many big tourist attractions and closing down the high-speed rail to China \cite{HKparkclose} \cite{HKsuspendrail}. Hence, after another week, the lockdown parameter was enabled from day 14 with a relatively high number of intra-city travellers and essential workers, resembling a semi-lockdown.

After three months, Hong Kong began easing its border control by giving quarantine exemptions to people who met certain conditions \cite{quarantineexemption}, and this is modelled in the simulation by increasing the number of randomly infected people from 2 per week to 10 per week after day 120. Simultaneously, Hong Kong began gradually easing its curfew by partially reopening public facilities \cite{HKreopen1} \cite{HKreopen2} and increasing the 8 people rule to 50. This is modelled in the simulation by disabling the lockdown and social distancing on day 126.

After approximately three weeks, the government tightened their borders again, including banning flights from India \cite{banindia}, which is replicated by reducing the randomly selected weekly infections from 10 to 2 again. Also, a semi-lockdown was enforced again, which is replicated in the model by re-enabling social distancing and total lockdown.

\subsubsection*{Hong Kong Validation Results}
\begin{figure*}[p]
    \includegraphics[scale=0.6,center]{{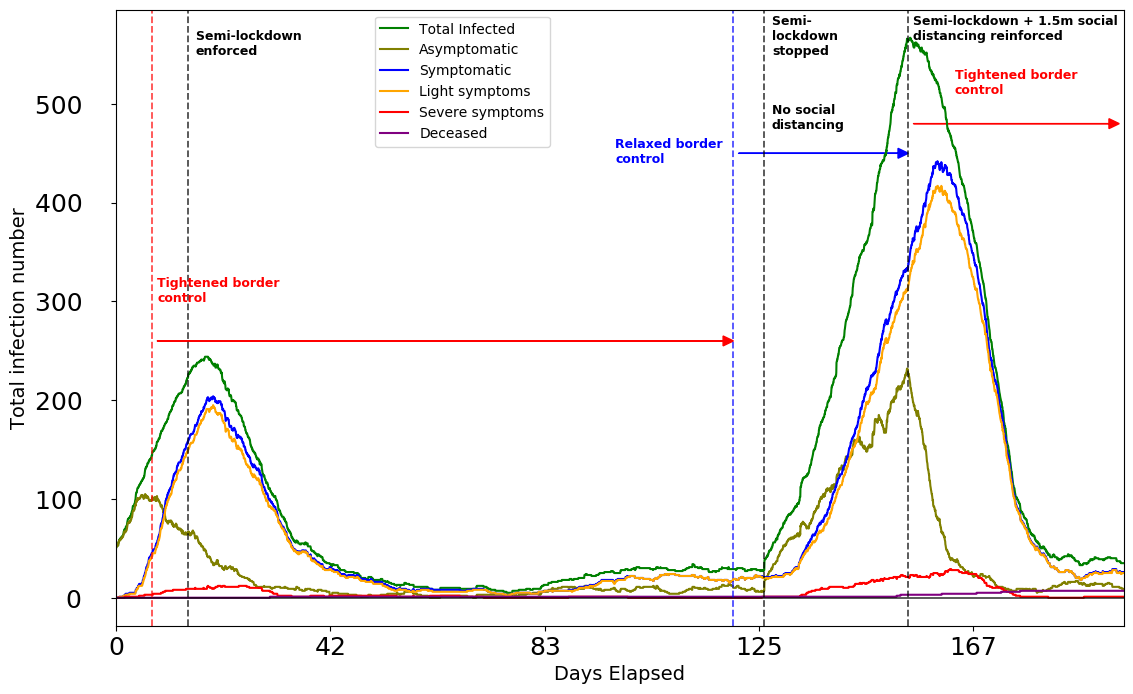}}
    \caption{Model results for Hong Kong}
    \label{hkfigure}
    \vspace{0.5cm}
    \includegraphics[scale=0.75,center]{{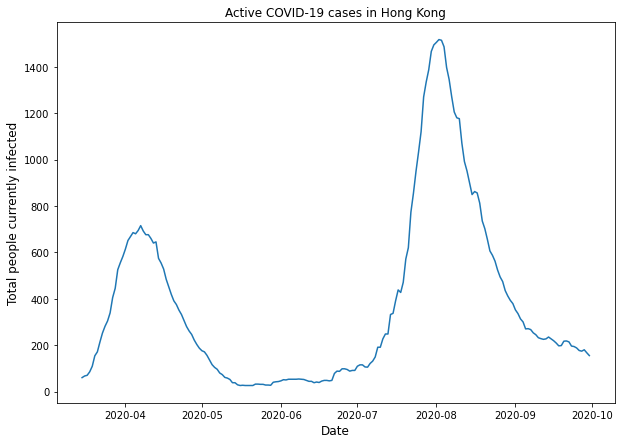}}
    \caption{Actual active cases in Hong Kong \cite{worldometerhk}} 
    \label{HKinfectionfigure}
\end{figure*}

Initially, the model infection numbers began rising at a relatively fast pace, but the tightening of borders and semi-lockdown were effective and infection numbers came to a halt at around day 20 and cases started to reduce again and were almost back to zero by approximately day 50, resembling the first wave and second wave of COVID-19 in Hong Kong, compared to the real-world data figure as seen in Figure \ref{HKinfectionfigure}.

From day 120, the increase of imported case, reduction of social distancing and disablement of semi-lockdown caused the numbers to rapidly spike to approximately twice as high as the initial spike, which also resembles the real-world data. 

From day 141, the tightening of borders, increase of social distancing and a second semi-lockdown quickly dampened again. Overall, the model produced an infection curve strongly resembling historical data in Hong Kong, as seen by the simulation plot in Figure \ref{hkfigure} and the real-world data plot in Figure \ref{HKinfectionfigure}. The raw number of infections was also significantly smaller than the infections in the UK and Italy, strongly resembling the real-world trend.

A comparison between the sampled normalised model results and real-world data of Hong Kong showed that the Pearson correlation coefficient is $0.40$ with a corresponding p-value of $3.65\times10^{-9}$, and the Spearman's rank correlation coefficient is $0.81$ with a corresponding p-value of $1.31\times10^{-48}$. Both correlation coefficients are fairly large and the p-value is extremely small, which is strong evidence of correlation. This suggests that the model can produce a fairly accurate replica of the pandemic in Hong Kong and successfully captured the trend. Any difference is possibly due to a change in the difference between model parameters and real-world parameters.

\begin{table}[!tb]
\centering
\begin{tabular}{|c|c|}
\hline
\textbf{Age Group} & \textbf{\% of Population} \\ \hline
0-9                & 0.0859      \\ \hline
10-19              & 0.0746      \\ \hline
20-29              & 0.1203      \\ \hline
30-39              & 0.1512      \\ \hline
40-49              & 0.1514      \\ \hline
50-59              & 0.1648      \\ \hline
60-69              & 0.1358      \\ \hline
70-79              & 0.0658      \\ \hline
80+                & 0.0501      \\ \hline
\end{tabular}
\caption{Age structure of Hong Kong's population} \cite{UNpopulation}
\label{HKpoptable}
\end{table}

\subsubsection{Italy}
The age distribution used for the simulation of Italy is provided in table \ref{ITpoptable}, calculated from data provided by the United Nations \cite{UNpopulation}.

\subsubsection*{Italy Validation Setup}
The simulation starts from the moment Italy received their initial case. Italy almost immediately stopped all the direct flights from China, whilst keeping all other flights open. Hence, the initial random number of infections per week is set to 25. Not many people are initially aware of the virus and many thought it was a common cold until awareness was raised via media \cite{italyunderestimate}. Hence, the initial symptomatic isolation rate is set at 30\%, and later raised to 60\% in the 3\textsuperscript{rd} day, and 90\% on the seventh day after people became increasingly aware of the virus.

As the virus quickly became out of control, Italy began closing some cities \cite{italytimeline}, which is modelled by enabling city confinement on the 7\textsuperscript{th} day. Schools also closed soon afterwards \cite{italytimeline}, which increases social distancing. Hence, this is modelled by increasing the social distancing metres to 0.5, also on the 7\textsuperscript{th} day. Note that although cities are confined, essential workers can still travel between cities.

The government then began strictly enforcing 1 metre of social distancing \cite{italysocialdistancing1m}, which is modelled by increasing the social distancing metres to 1 on the 10\textsuperscript{th} day.

Although the rate of infections has reduced, the curve was still rising, and Italy further increased their safety measures by enforcing mobility restrictions, followed by a total lockdown \cite{italytimeline}. This is implemented in the model by increasing the social distancing metres to 1.5 on day 12 and enabling total lockdown as well as decreasing the random infections to 6 per week on day 15 respectively.

This gradually stopped the increase of cases, and the curve quickly dropped, similar to the real data. After an extended period of lockdown, the government began reopening public facilities and relaxing the lockdown rules \cite{italyeaselockdown}, which is modelled by disabling the total lockdown and reducing the social distancing metres to 1. This caused the rate of reduction to slow down. Eventually, Italy partially reopened its borders to some neighbouring European countries \cite{italyliftrestriction}, which is modelled in the simulation by increasing the random weekly infection to 15 per week.

\subsubsection*{Italy Validation Results}
\begin{figure*}[p]
    \includegraphics[scale=0.6,center]{{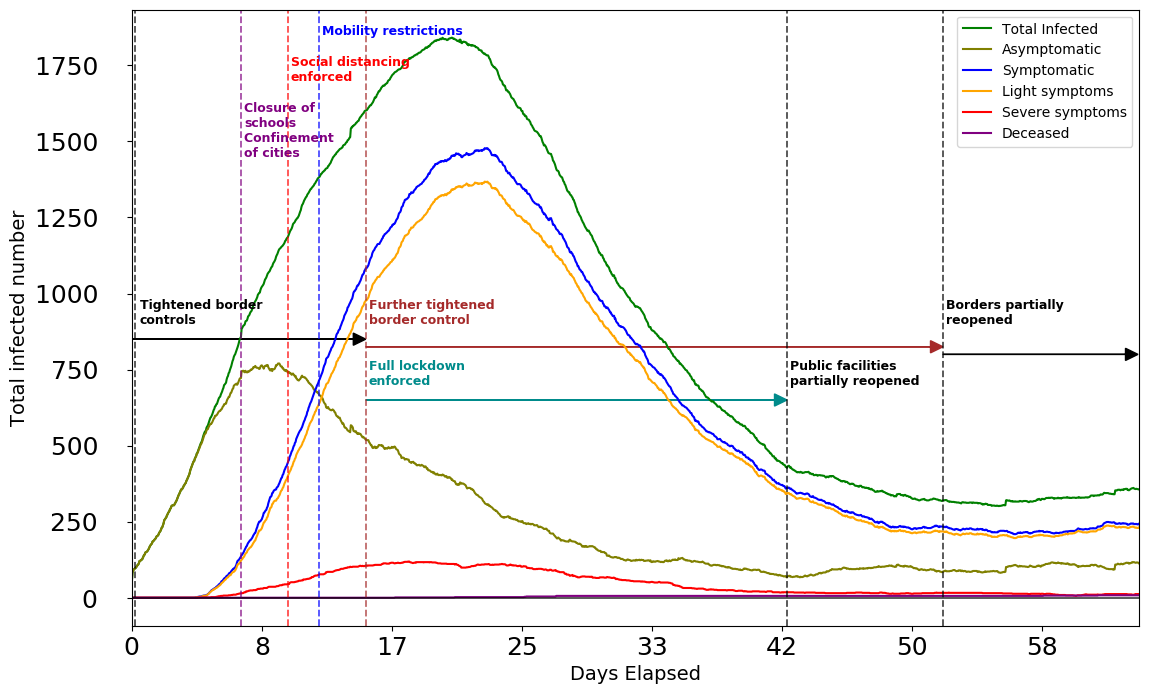}}
    \caption{Model results for Italy}
    \label{itfigure}
    \vspace{0.5cm}
    \includegraphics[scale=0.75,center]{{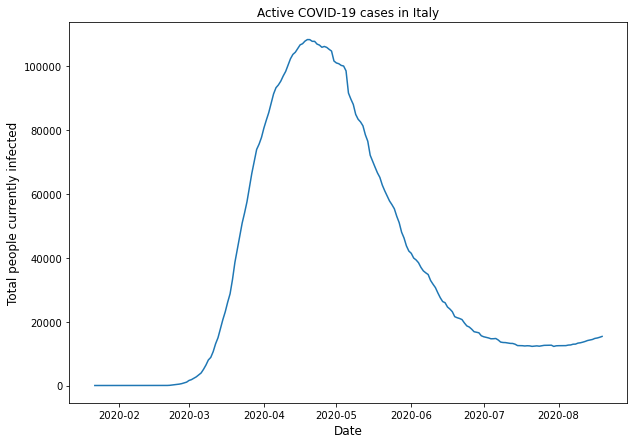}}
    \caption{Actual active cases in Italy \cite{worldometeritaly}} 
    \label{ITinfectionfigure}
\end{figure*}

With minimal NPI's enforced, the infection numbers began rapidly rising for the first week until social distancing measures and confinement were introduced. However, from Figure \ref{itfigure}, it can be seen that the cases were still rapidly increasing, which indicated that the preventative measures were not sufficient.

It was not until the enforcement of a full lockdown on day 15 where the infections stopped increasing, and the curve quickly dropped, which resembled the historical data, as ween in figure \ref{ITinfectionfigure}.

However, although the infection number was dropping, it was still far from zero when the public facilities partially reopened on day 42 of the model simulation. After the public facilities reopened, the infection numbers still decreased, but at a slower rate. After the reopening of borders, the infection curve stopped decreasing and began slowly rising again, which accurately reflected the historical trend where the infection numbers began rising towards the end of August, as seen in Figure \ref{ITinfectionfigure}. Based on these model results, it indicated that a new wave may be potentially coming, which was correct, as a second wave soon arrived, according to historical data \cite{johnshopkins}. Overall, there was a large resemblance between the model results and the historical data, as seen in Figures \ref{itfigure} and \ref{ITinfectionfigure}.

A comparison between the sampled normalised model results and real-world data of Italy showed that the Pearson correlation coefficient is $0.66$ with a corresponding p-value of $7.89\times10^{-28}$, and the Spearman's rank correlation coefficient is $0.73$ with a corresponding p-value of $4.38\times10^{-36}$. These correlation coefficients are both relatively high and the p-values are very close to zero, which is evidence that the model can produce a reasonably accurate replica of the pandemic in Italy and successfully captured the trend.

\begin{table}[!tb]
\centering
\begin{tabular}{|c|c|}
\hline
\textbf{Age Group} & \textbf{\% of Population} \\ \hline
0-9                & 0.0843      \\ \hline
10-19              & 0.0948      \\ \hline
20-29              & 0.1013      \\ \hline
30-39              & 0.1173      \\ \hline
40-49              & 0.1524      \\ \hline
50-59              & 0.1561      \\ \hline
60-69              & 0.1221      \\ \hline
70-79              & 0.0980      \\ \hline
80+                & 0.0738      \\ \hline
\end{tabular}
\caption{Age structure of Italy's population} \cite{UNpopulation}
\label{ITpoptable}
\end{table}

\subsubsection{United Kingdom}
The age distribution used for the simulation of the UK is provided in table \ref{UKpoptable}, calculated from data provided by the United Nations \cite{UNpopulation}.

\subsubsection*{UK Validation Setup}
Initially, the United Kingdom approached the pandemic with the herd immunity strategy \cite{borisherd}, where minimal countermeasures were used except for raising awareness of the virus and encouraging people with symptoms to self-isolate at home. Hence, the experiment was set up with no social distancing as well as a mask usage rate of only 5\% \cite{statistamask}. Initially, there was very little awareness of the virus and many people still went outside despite feeling symptoms, thinking it was a common flu. Hence the initial symptomatic isolation rate was set to 30\%. The UK airport was fully open during the start of COVID-19, hence to model imported COVID-19 cases, 40 out of 10000 people are randomly infected every week initially.

For the first 9 days, no significant changes were made in the simulation model except that the symptomatic isolation rate was increased to 60\% on day 3, and then to 90\% on day 7, as people become more aware of COVID-19.

Eventually, the government abandoned the herd immunity strategy and began partially closing public facilities and encouraging 2 metres of social distancing. To model this in the simulation, the social distancing metres was increased to 2.

The UK then went into full lockdown and further tightened its borders, which was modelled in the simulation by enforcing a total lockdown and reducing the number of random infections from 40 per week to 6, from day 15. Approximately 2 weeks after the peak \cite{ukeaselockdown}, the UK began partially easing its lockdown rules and more people began going outside. This is modelled by decreasing the social distancing metres to 1.5 from day 28. Border controls were eventually more lenient as neighbouring countries saw a drop in COVID-19 cases, and this is modelled by increasing the random infections from 6 to 18, from day 38. Eventually, the UK dropped the lockdown rule and reopened most of its facilities, which is modelled by decreasing the social distancing metres to 0.5 and disabling the total lockdown.

As the UK stopped releasing daily recovered cases data after April 13\textsuperscript{th} 2020 \cite{johnshopkins}, the active cases could not be calculated and as a result, the active cases could not be plotted. Hence, the validation numbers are compared to the number of real daily cases, which also possesses a similar distribution.

\subsubsection*{UK Validation Results}
\begin{figure*}[!tb]
    \includegraphics[scale=0.6,center]{{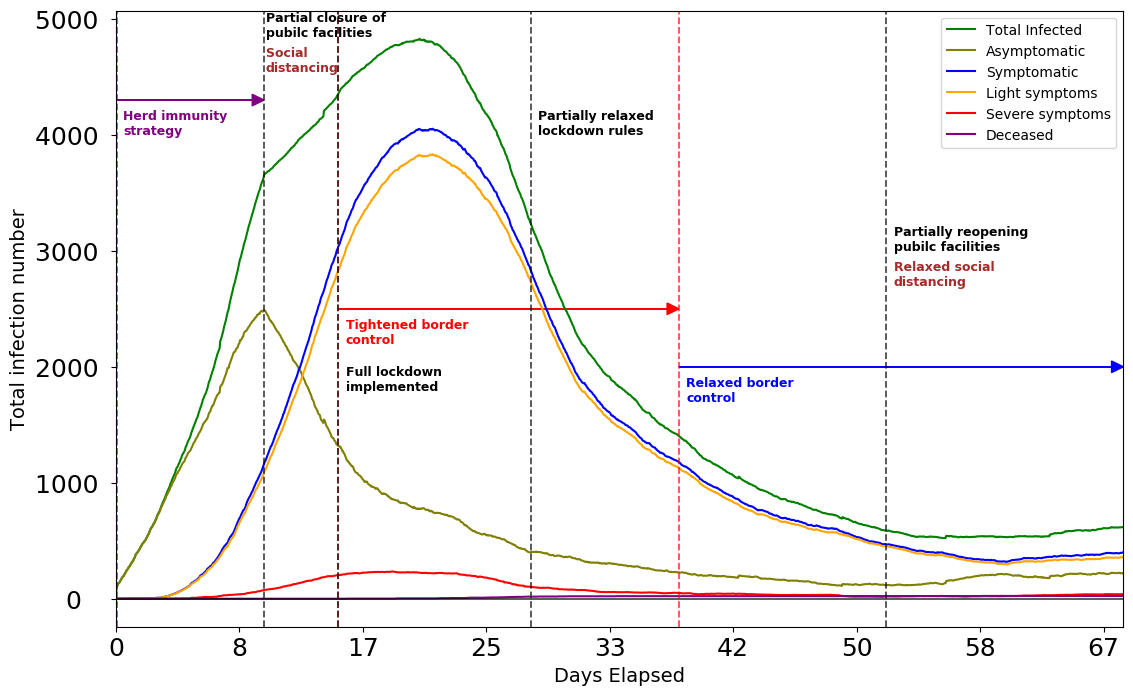}}
    \caption{Model results for the UK}
    \label{ukfigure}
\end{figure*}

As a result of the herd immunity strategy, the number of cases rapidly increased for the first 9 days. After the UK changed the strategy and increased social distancing and put closed facilities. This had an effect on slowing down the rate of increase of infections, as visualised by the green curve in Figure \ref{ukfigure}, followed by the blue and orange curve due to the symptomatic delay. However, it can be seen that the curve was still increasing, indicating that these NPI's were not strong enough to contain the virus.

It was not until the enforcement of a full lockdown and tightening of border control where the active cases stopped increasing. The number of active cases then began to slowly decrease after day 22.

The relaxing of lockdown rules on day 28 cause the rate of decrease of cases to slow down and the relaxing of border control further reduced the rate decrease. Eventually, the partial reopening of facilities caused the number of infections to gradually rise again, resembling the historical trend in Figure \ref{countriescomparison}, suggesting a new wave is approaching, which is correct according to historical data \cite{johnshopkins}. The overall distribution of the model infection curve resembles the historical data fairly strongly.

A comparison between the sampled normalised model results and real-world data of the UK showed that the Pearson correlation coefficient is $0.92$ with a corresponding p-value of $5.09\times10^{-57}$, and the Spearman's rank correlation coefficient is $0.95$ with a corresponding p-value of $2.46\times10^{-71}$. Both correlation coefficients are very close to 1 and the p-values are very close to zero, which indicates that the model can produce a very accurate replica of the pandemic in the United Kingdom and successfully captured the trend.

\begin{table}[!tb]
\centering
\begin{tabular}{|c|c|}
\hline
\textbf{Age Group} & \textbf{\% of Population} \\ \hline
0-9                & 0.1194      \\ \hline
10-19              & 0.1121      \\ \hline
20-29              & 0.1278      \\ \hline
30-39              & 0.1363      \\ \hline
40-49              & 0.1277      \\ \hline
50-59              & 0.1353      \\ \hline
60-69              & 0.1067      \\ \hline
70-79              & 0.0840      \\ \hline
80+                & 0.0506      \\ \hline
\end{tabular}
\caption{Age structure of UK's population} \cite{UNpopulation}
\label{UKpoptable}
\end{table}

\subsection{Discussion}
Although the results from the univariate analysis suggested that a lockdown was the quickest way to instantly reduce the infection numbers in the shortest time out of the four safety measures, the results from the multivariate analysis showed that social distancing and usage of surgical masks have a more significant effect on reducing infection numbers, as they have relatively higher values of $\mu\mbox{*}$ and $\sigma$.

\section{Conclusion and Future Work}
Whilst the resulting model can produce a reasonably fair estimation of the trend of COVID-19 as well as determine the effectiveness of certain preventative measures, for it to be used by policy makers to make key decisions, the model needs to be further refurbished and more complex, i.e. take into account more real-life factors. Instead, due to its ease of use and flexibility, it can be used as a tool and source for policy makers to back up their various hypotheses in combination with other research and models, and to visualise how certain interventions will affect the infection trend.

One could extend this model in the future by adding features such as small venues, schools and transport, which will allow the model to make even more realistic predictions. One can also continue to monitor trends from different countries and observe how the findings of the Morris Elementary Effects method relate to the trend.

In conclusion, the results from the univariate and multivariate analysis have strongly suggested that there would be no need to enforce a lockdown at all if a sufficient proportion of the population followed the social distancing and mask usage guidelines, as the peak infection number would be sufficiently controlled and relatively low.

\ifCLASSOPTIONcaptionsoff
  \newpage
\fi

\section*{Acknowledgements}
This study was based on the findings of the lead author's dissertation project which was completed during his time at the University of Warwick, although he is now working as a Data Scientist for the Chinese University of Hong Kong. 

FM is supported by the PathLAKE digital pathology consortium which is funded from the Data to Early Diagnosis and Precision Medicine strand of the government's Industrial Strategy Challenge Fund, managed and delivered by UK Research and Innovation (UKRI).

\bibliographystyle{unsrt}
\bibliography{ref.bib}

\end{document}